\documentclass[sigconf]{acmart}
\usepackage{mathtools}
\usepackage{threeparttable}
\usepackage{multirow}

\AtBeginDocument{%
  \providecommand\BibTeX{{%
    \normalfont B\kern-0.5em{\scshape i\kern-0.25em b}\kern-0.8em\TeX}}}

\begin{document}
\title{Psychometric Analysis and Coupling of Emotions Between State Bulletins and Twitter in India during COVID-19 Infodemic}

\author{Baani Leen Kaur Jolly}
\authornote{These authors contributed equally to this research.}
\email{baani16234@iiitd.ac.in}
\affiliation{%
  \institution{Indraprastha Institute Information Technology, Delhi}
}
\author{Palash Aggrawal}
\authornotemark[1]
\email{palash16064@iiitd.ac.in}

\affiliation{%
  \institution{Indraprastha Institute Information Technology, Delhi}
%   \city{New Delhi}
%   \state{India}
}

\author{Amogh Gulati}
\authornotemark[1]
\email{amogh17019@iiitd.ac.in}

\affiliation{%
  \institution{Indraprastha Institute Information Technology, Delhi}
}

\author{Amarjit Singh Sethi}
\email{ajss.gndu@gmail.com}

\affiliation{%
  \institution{Guru Nanak Dev University, Amritsar}
}

\author{Ponnurangam Kumaraguru}
\email{pk@iiitd.ac.in}
\affiliation{%
  \institution{Indraprastha Institute Information Technology, Delhi}
}
\author{Tavpritesh Sethi}
\email{tavpriteshsethi@iiitd.ac.in}
\affiliation{%
  \institution{Indraprastha Institute Information Technology, Delhi}
}

\begin{abstract}
COVID-19 infodemic has been spreading faster than the pandemic itself. The  misinformation riding upon the infodemic wave poses a major threat to people's health and governance systems. Since social Media is the largest source of information, managing the infodemic not only requires mitigating of misinformation but also an early understanding of psychological patterns resulting from it. During the COVID-19 crisis, Twitter alone has seen a sharp 45\% increase in  the usage of its curated events page, and a 30\% increase in its Direct messaging usage, since March 6th 2020.\footnote{\url{https://blog.twitter.com/en_us/topics/company/2020/An-update-on-our-continuity-strategy-during-COVID-19.html}}\\
In this study, we analyze the psychometric impact and coupling of the COVID-19 infodemic with the official bulletins related to COVID-19 at the national and state level in India. We look at these two sources with a psycho-linguistic lens of emotions and quantified the extent and coupling between the two. We modified \textit{Empath}, a deep skip-gram based open sourced lexicon builder for effective capture of health related emotions. We were then able to capture the time-evolution of health-related emotions in social media and official bulletins. An analysis of lead-lag relationships between the time series of extracted emotions from official bulletins and social media using Granger's causality showed that state bulletins were leading the social media for some emotions such as Medical Emergency. Further insights that are potentially relevant for the policymaker and the communicators actively engaged in mitigating misinformation are also discussed. Our paper also introduces CoronaIndiaDataset\footnote{\url{http://precog.iiitd.edu.in/resources.html}}, the first social media based COVID-19 dataset at national and state levels from India with over 5.6 million national and 2.6 million state level tweets. Finally, we present our findings as \href{http://covibes.tavlab.iiitd.edu.in/ }{COVibes}\footnote{\url{http://covibes.tavlab.iiitd.edu.in/ }}, an interactive web application capturing psychometric insights captured upon the CoronaIndiaDataset, both at a national and state level.
\end{abstract}

\maketitle

\section{Introduction}
"We’re not just fighting an epidemic; We’re fighting an \textit{infodemic}". These words spoken by the WHO Director-General at the Munich Security Conference on 15 February 2020\footnote{\url{https://www.who.int/dg/speeches/detail/munich-security-conference}}, sums up the challenges faced by our society due to COVID-19. Infodemic comes from the root words \textit{information} and \textit{epidemic}. It refers to an excessive amount of information which is made publicly available, consisting of both accurate and not so accurate information, which makes it harder to find reliable, trustworthy and accurate guidance when needed.\footnote{\url{https://www.who.int/docs/default-source/coronaviruse/situation-reports/20200415-sitrep-86-covid-19.pdf}}.\\
Social media is one of the most popular media for the diffusion of information during an infodemic. Twitter, a micro-blogging site, is one of the most widely used social media platforms and it has seen a sharp 45\% increase in the usage of its curated events page, and a 30\% increase in its Direct messaging usage, since March 6th 2020, during the COVID-19 emergency.\footnote{\url{https://blog.twitter.com/en_us/topics/company/2020/An-update-on-our-continuity-strategy-during-COVID-19.html}}. Social media platforms are not only easily accessible and have a global reach, they provide a virtual space to the users distant from their real world. Due to this, they provide a medium to help people discuss many taboo topics which they might not in their real social circles, like mental healthcare, domestic violence, sexual assault etc. This provides an opportunity to use data mining \& natural language processing techniques to analyse the web data from the standpoint of psychology.\\ 

The recent outbreak of COVID-19 (COrona VIrus Disease 2019) has the world in its grips. With exponentially rising cases, WHO declared it as a global pandemic on March 11, 2020. Following health advisories and the virus rapidly spreading, most countries have declared national emergencies, closed borders and restricted public movement. Most countries in the world have gone into national lockdown to contain the pandemic.

Naturally, such a strong event's impact has affected the daily lives of people all over the world, something which has also impacted people's social media usage. In recent times of social distancing, social media has  become a popular platform for people to express their thoughts \& opinions. Social Media is not only being used by people to express how their lives have been affected and appeal to everyone to understand the importance of the situation, but trending hashtags like \#CoronaOutbreak \#COVID19, are used by authorities disseminate important information and health advisories.\\

A lot of research has previously been done on social media analysis related to  pandemics. Ritterman et al. \cite{Ritterman2009} showed how prediction market models and social media analysis can be used to model public sentiment on the spread of a pandemic. Signorini et al. \cite{signorini2011use} examined Twitter based-information to track the swiftly-evolving public sentiment regarding Swine Flu, in 2011 as well as correlate the H1N1 related activity to accurately track reported disease levels in the US. Jain et al.\cite{jain2015effective} used Twitter as a surveillance system to track the spread of 2015 H1N1 pandemic in India as well as the general public awareness towards it.
In this paper, we examine the use of social media during this ongoing time of n-CoV2019 pandemic. As the impact of this pandemic is growing at an exponential rate, we try to model this rapidly evolving public sentiment and dig deeper to understand how it changes daily, using time-series based analysis. To this end, we have curated dataset of more than 5.6 million tweets and retweets, specific to India. This dataset was collected using two approaches - Content-Based and Location-Based queries, as explained in Section \ref{sec:dataset}. We model the public sentiment using sentiment analysis and the Empath psycho-linguistic features.

\section{Literature Survey}
Prior work done on understanding the public sentiment during the Ebola medical crisis by Lazard et al. \cite{lazard2015detecting} detected the ongoing themes in the ongoing discourse on the live Twitter chat by Centers for Disease Control and Prevention. Wong et al. \cite{wong2017local} further studied the tweets by the local health departments related to Ebola. The work done by Kim et al. \cite{kim2016topic} presents a topic-based sentiment analysis of the Ebola virus on twitter and in the news.

With the world's efforts focused on battling COVID-19, research efforts from many fields including online social media \& textual analysis has vigorously started in this direction. Chen et al. \cite{Chen2020} provide the first Twitter dataset by collecting tweets related to \#COVID19. This is an ongoing collection, started from January 22, 2019. Haouari et al. \cite{haouari2020ArCov-19} also present ArCov-19, a large Arabic Twitter dataset collected from January 27,2019 to March 31, 2019.

During this infodemic, Zhao et al.\cite{Zhao2020} analyse the attention of Chinese public to COVID-19 by analysing search trends on Sina Microblog and evaluating public opinion through word frequency and sentiment analysis. Cinelli et al. \cite{cinelli2020covid} provide insights into the evolution of the global COVID-19 discourse on Twitter, Instagram, Reddit, YouTube and Gab.
Alhajji et al. \cite{Alhajji2019} analysed public sentiment of Saudi Arabia by collecting upto 20,000 tweets within 48 hours of key events in Saudi Arabia's timeline, using transfer learning to do sentiment analysis. Kayes et al. \cite{Kayes2020} attempt to measure community acceptance of social distancing in Australia, reporting that majority of tweets were in favour of social distancing. Li et al. \cite{Li2020} try to extract psychological profiles of active Weibo users during the time of COVID-19 spread in China to analyse linguistic, emotional and cognitive indicators.\\

Analysing user behavior and response can provide a critical understanding of what policies and decision worked. Hou et al. \cite{Hou2020} assess the public's response to the situation and the government guidelines in terms of attention, risk perception, emotional and behavioral response by analysing search trends, shopping trends and blog posts on popular Chinese services. Li et al. \cite{Li2020b} analyse how information regarding COVID-19 was disseminating, suggesting useful insights into the need for information. Work by Schild et al. \cite{Schild2020} shows how the current pandemic situation has caused an unfortunate rise in Sinophobic behavior on the web.

\section{Dataset}
\label{sec:dataset}
We curated the \textit{CoronaIndiaDataset(CID)}, collected using Twitter's official Tweepy API, from 1st March to 27th April 2020. CoronaIndiaDataset contains 56,24,066 tweets talking about COVID-19 in Indian settings as well as 26,55,503 state-specific tweets. Table \ref{tab:CID_Stats} shows the state-wise frequency of tweets available in our dataset.

\subsection{Twitter Data Collection}% Methodology}
 The data was collected using 2 separate approaches in parallel -- \textit{Content Based Query} and \textit{Location Based Query}.
 
\subsubsection{Content-based Query:}
To collect the relevant Twitter data, we explored the trending and most popular hashtags for each of the Indian states and manually curated the list of hashtags related to COVID-19. We also went through several n-CoV2019 related tweets manually to find and subsequently mined the most popular hashtags(list given in appendix) related to the same, which may not be trending. Further, in order to automate the collection of relevant tweets, we tried to formulate generic queries like `corona <state>', etc. and collected the state-wise n-CoV2019 related twitter data using the same.\\
This approach focuses on getting all tweets which are talking about COVID-19 in the context of India or the Indian states. Multiple queries were built by joining terms related to COVID-19 with the name or common aliases of the region and data was collected from 1st March to 23rd April 2020. Some of the terms used were - `corona', `covid19', `coronavirus', `lockdown', etc. In addition, popular hashtags like \#coronavirusin<region>, \#coronain<region>, \#corona<region>, \#<region>fightscorona were used, where `<region>` is replaced by the name or popular alias of India or various states. Examples of a popular alias are `Orissa' for Odisha, `TN' for Tamil Nadu, `UP' for Uttar Pradesh, or spelling mistakes like `chatisgarh' for Chhattisgarh.
% , as shown in Figure \ref{fig:twitter-india-availability}. 
% Moreover, to collect the state-wise Twitter data, we explored the trending hashtags for each of the Indian states and manually curated the list of trending hashtags for the same. We used these in addition to the generic query `corona <state>' and collected the state-wise Corona related twitter data using the same. We also went through several corona-virus related tweets manually to find and subsequently mine the most popular hashtags related to the same, which may not be trending.\\

\begin{table*}[t]
\centering
\caption{CoronaIndiaDataset Statistics.}
\label{tab:CID_Stats}
\begin{tabular}{|l|r||l|r||l|r|}
\hline
Region &    Tweets Available & Region &    Tweets Available & Region &    Tweets Available \\
\hline
\textbf{India             } &  56,24,066 & \textbf{Chhattisgarh      } &     32,058 & \textbf{JnK               } &     15,923 \\
\textbf{Delhi             } &   5,57,458 & \textbf{Madhya Pradesh    } &     82,290 & \textbf{Jharkhand         } &     24,276 \\
\textbf{Karnataka         } &   2,09,582 & \textbf{Maharashtra       } &   3,89,982 & \textbf{Himachal Pradesh  } &     13,017 \\
\textbf{Tamil Nadu        } &   2,08,608 & \textbf{Uttar Pradesh     } &   2,33,507 & \textbf{Assam             } &     16,707 \\
\textbf{Telangana         } &   1,75,444 & \textbf{Goa               } &     11,943 & \textbf{Meghalaya         } &      2,641 \\
\textbf{Haryana           } &     95,197 & \textbf{Rajasthan         } &   1,18,571 & \textbf{Mizoram           } &      1,547 \\
\textbf{West Bengal       } &     93,861 & \textbf{Gujarat           } &     97,478 & \textbf{Arunanchal Pradesh} &      1,388 \\
\textbf{Odisha            } &     50,704 & \textbf{Bihar             } &     93,623 & \textbf{Manipur           } &      2,796 \\
\textbf{Kerala            } &     47,644 & \textbf{Uttarakhand       } &     17,992 & \textbf{Tripura           } &      4,078 \\
\textbf{Punjab            } &     43,426 & \textbf{Sikkim            } &      1,126 & \textbf{Nagaland          } &      3,013 \\
\textbf{Andhra Pradesh    } &     18,759 & & & & \\
\hline
\end{tabular}
\end{table*}

\subsubsection{Location-based Query:}
Tweets are collected for globally trending COVID-19 related hashtags, and then filtering tweets based on the User Location. Tweets for the following hashtags and keywords are collected - \#COVID19, `CoronaVirusUpdates', `coronavirus', `corona virus outbreak', `corona wuhan', \#Coronavirus, \#NCOV19, \#CoronavirusOutbreak, \#coronaviruschina, \#coronavirus, `COVID19' from 14th March to 27th April 2020. This resulted in a collection of a total of 12 million tweets from all over the world. We also create a list of location filters for various states. These are the state names, aliases (as explained above) and name of popular cities in those states. Using these, we first filter out all tweets having `india' in their user location, and then sort them based on keyword matches of tokens in the user location  with the above list. The User Location was lowercased before matching.

\begin{figure*}[ht!]
	\centering
	\includegraphics[scale=0.40]{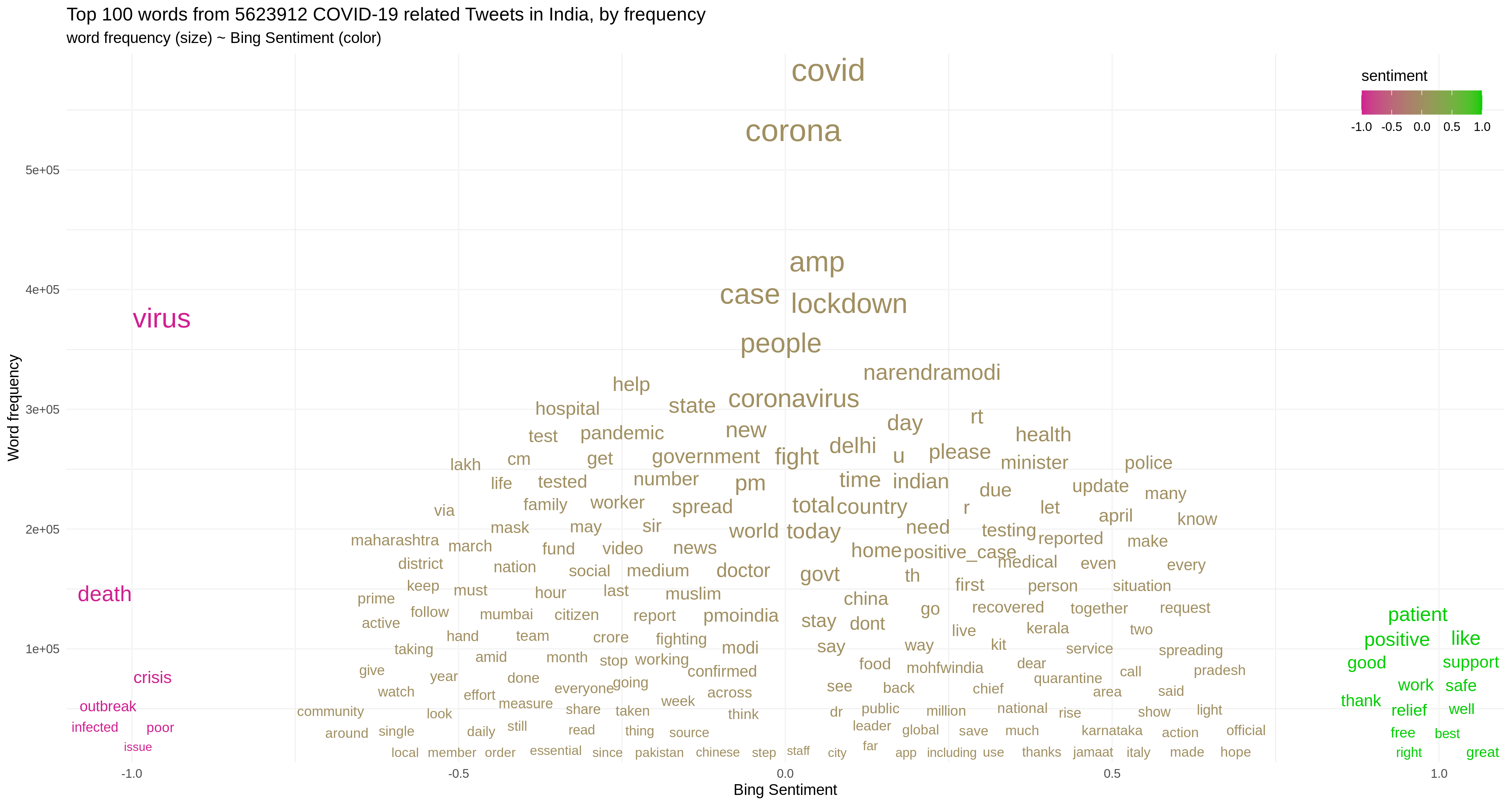}
	\caption{Chatterplot for COVID-19 India-Related Tweets. The X axis is the 
% 	AFINN-111 
	Bing
	Sentiment and Y axis is the frequency.}
	\label{fig:twitter-india-wordcloud}
\end{figure*}

We analysed using a chatterplot, the most frequent occurrences in our dataset, as shown in Figure \ref{fig:twitter-india-wordcloud}. The plot shows the top 200 words, arranged by their frequency and 
% AFINN sentiments \cite{nielsen11}. 
Bing Sentiments \cite{liu2012sentiment}. +1 is positive, -1 is negative and the group of words in the middle have no sentiment value associated. In this plot, we have removed the outlier 'India' for a better representation of other terms. 

\begin{figure}[ht!]
	\centering
	\includegraphics[scale=0.5]{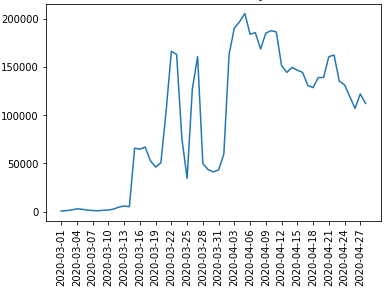}\\
	\caption{Frequency of tweets available in our Dataset.}
	\label{fig:twitter-india-availability}
\end{figure}

\subsection{Indian State Government n-CoV2019 related bulletins}
\label{govt-bulletins}
DataMeet Community has curated a database of COVID-19 related government bulletins from Indian States \footnote{http://projects.datameet.org/covid19/} \footnote{https://github.com/datameet/covid-19-indian-state-reports}. These bulletins have statistics about COVID-19 cases in the state, government's response to them, advisories and other useful information. We have analysed all the reports that are in English language and belong to the states of Delhi, West Bengal, Punjab, Tamil Nadu, Odisha and Kerala. 

% \begin{figure}
%     \centering
%     \includegraphics{}
%     \caption{Caption}
%     \label{fig:my_label}
% \end{figure}

\section{Methodology}

% \section{Methodology}
\subsection{Preprocessing Data}
We followed the below steps to pre-process the data and reduce the noise:
\begin{enumerate}
\item \textit{Lowercasing}
\item \textit{Tokenisation}
\item \textit{Links Removal}
\item \textit{Removal of all alphanumeric characters, allowing only standard English alphabets}
\item \textit{Stop Words Removal}:\\
Performed Stop Words removal, using NLTK, for English and 12 other languages like French, Spanish, German, which share the same script with English.
\item \textit{Lemmatisation}:\\
We used Wordnet Lemmatiser instead of Porter Stemmer, since the latter leads to the reduction of words to a form wherein they are no longer real words, however, the former ensures that each word is reduced to a real word in English dictionary. For example, `studies' and `studying' get converted to `studi' and `study' by Porter Stemmer, while a Lemmatiser matches both of them to a common lemma `study'.
\end{enumerate}

\subsection{Quantitative Empath Analysis}
\label{subsec:empath-def}
Empath \cite{Empath-Fast2016} is an open vocabulary based tool to generate and validate lexical categories. It is based on deep skip-gram model to draw correlation between many words and phrases starting from a small set of seed words. It has some inbuilt categories, including emotions, which can be used to identify the emotion associated with a text. \\

Empath provides 3 types of datasets to build the lexicon from - `reddit' (social media), `nytimes' (news articles) and `fiction', and models a category by finding the words closest to the "seed words" of that category. But, the text data used for either of them is outdated and does not have enough information about the language used in the current scenario. Preliminary analysis using the Empath library showed that the current lexicon was inadequate to properly analyse the current situation. A case in point would be that the word `positive' had a connotation with positive emotion in the Empath categories, however, in the COVID-19 scenario, it was often used in the context of `tested positive',  which by itself was neither a positive nor a negative emotion, and rather hinted at the activity of testing positive for COVID-19

To rectify this, we manually examined the most frequent unigrams and bigrams in the collected data as well as some common bigrams in the given context which may be classified incorrectly, and manually annotated them into the most relevant categories or created new categories to help better analyse the emotional content of the tweets. Some important modifications are shown in Table \ref{tab:empath-mod}.

\begin{table}[h]
\centering
\caption{Important modifications to Empath categories.}
\label{tab:empath-mod}
\begin{tabular}{|p{2.5cm}|p{4cm}|} 
\hline
\textbf{Empath Category}    & \textbf{Tokens added (Lemmatized)}                                                        \\ 
\hline
medical emergency  & `case', `positive', `positive case', `test\_positive', `pandemic', `lockdown', `spread'   \\ 
\hline
health              & `test\_positive', `test\_negative'                                               \\ 
\hline
healing             & `vaccine'                                                                        \\ 
\hline
government & `cm', `pm', `prime\_minister', `minister', `govt'                                 \\ 
\hline
movement            & `socialdistancing', `social\_distancing', `awareness'                            \\ 
\hline
fight, war          & `eradicate', `contain', `overcome', `prevent'                                              \\ 
\hline
business            & `startup'                                                                        \\ 
\hline
\end{tabular}
\end{table}

% The main advantage Empath holds over other similar tools like LIWC [CN], is the larger size and ad-hoc custom category creation\cite{Empath-Fast2016}, which is important to properly analyse this particular situation. 
We analyse Empath scores of emotions related to Positive Sentiment, Negative Sentiment, Country and Government, the pandemic caused by COVID-19 and the fight against COVID-19. Details of the specific categories used can be found in Table \ref{tab:empath-cats}.

%'help', `medical emergency', `health', `hygiene', `fear', `death', `negative emotion', `sadness', `nervousness', `confusion', `war', `fight', `healing', `movement', `positive emotion', `optimism', `sympathy', `family', `government', `modi', `economics', `business', `occupation'. 

\begin{table}[h]
\centering
\caption{Empath Categories.}
\label{tab:empath-cats}
\setlength{\tabcolsep}{10pt} % Default value: 6pt
\renewcommand{\arraystretch}{1.5} % Default value: 1
\begin{tabular}{|p{2.5cm}|p{4.5cm}|}
\hline
\textbf{Domain} & \textbf{Empath Categories}\\ 
\hline
Positive Sentiment   & `help', `healing', `positive emotion', `optimism', `sympathy'    \\
Negative Sentiment   & `fear', `negative emotion', `sadness', `nervousness', `confusion'  \\ 
\hline
Country            & `government', `economics', `business', `occupation'                  \\
Pandemic Situation & `medical emergency', `health', `hygiene'                                    \\
Fighting n-CoV2019    & `war', `fight', `movement'      \\
\hline
\end{tabular}
\end{table}

% %%%%%%%%%%%%%%%%%%% FP-Control EMPATH GRAPHS %%%%%%%%%%%%%%%%%%%%
\begin{figure}[ht!]
	\centering
	\begin{tabular}{ccc}
		\includegraphics[scale=0.3]{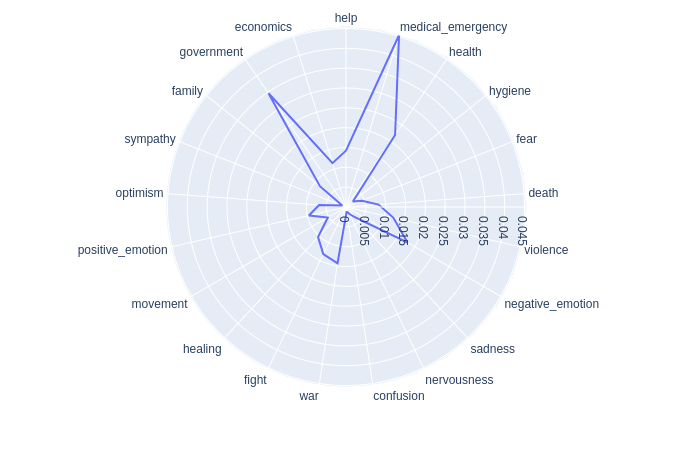} \\
	\end{tabular}
	\caption{Empath Analysis for Indian Twitter: \textit{medical\_emergency} \& \textit{government} are the most frequently talked about Empath categories.}
	\label{fig:empath-india}
\end{figure}

\begin{figure*}[ht!]
	\centering
	\begin{tabular}{ccc}
		\includegraphics[scale=0.25]{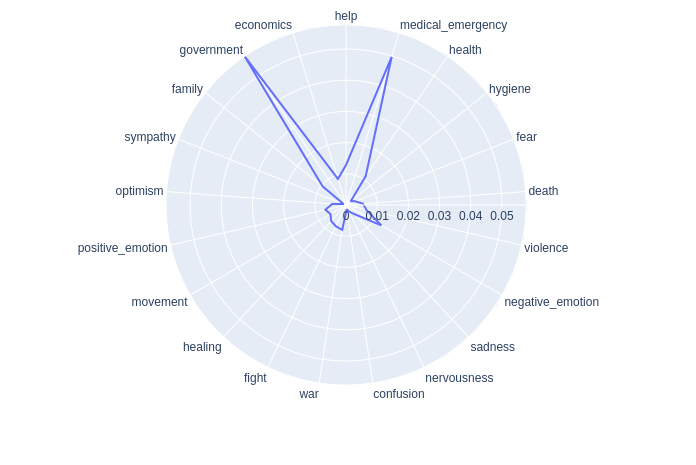} & 
		\includegraphics[scale=0.25]{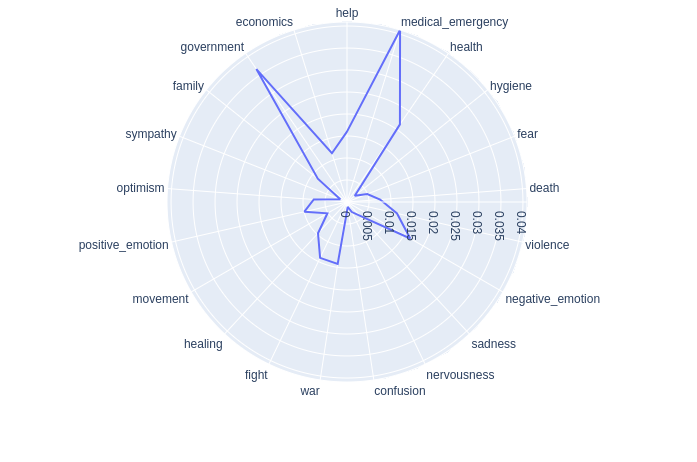} &
		\includegraphics[scale=0.25]{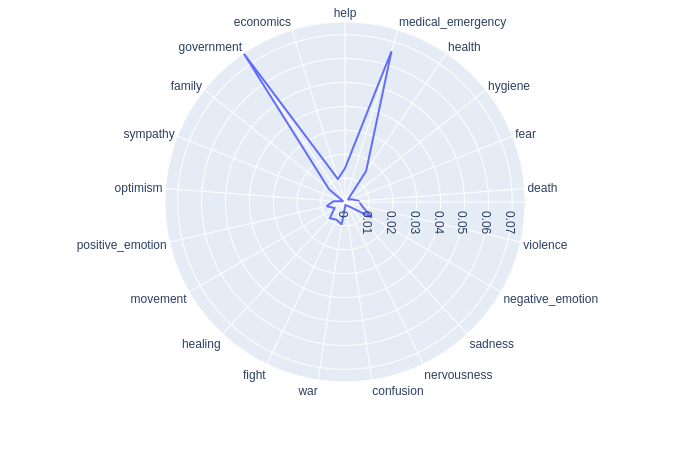} \\
		(a) Delhi & (b) West Bengal & (c) Punjab\\
		\includegraphics[scale=0.25]{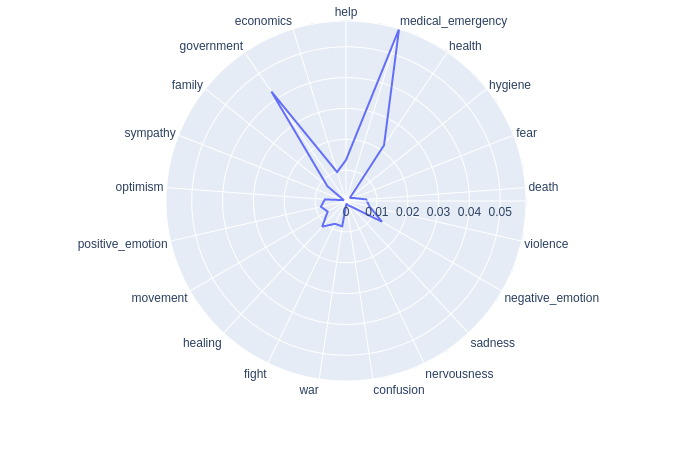} &
		\includegraphics[scale=0.25]{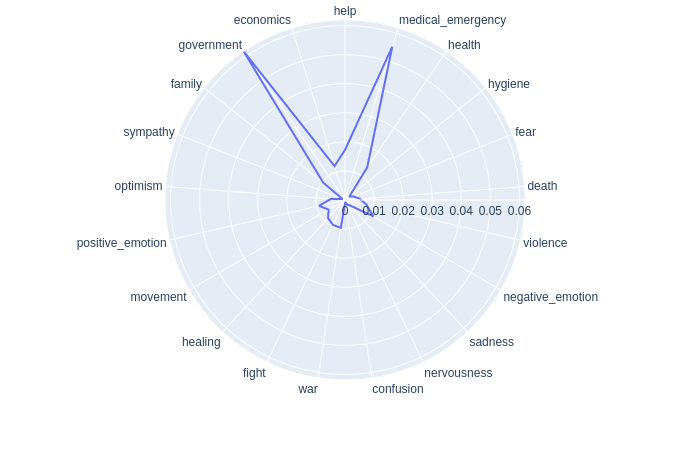} &
		\includegraphics[scale=0.25]{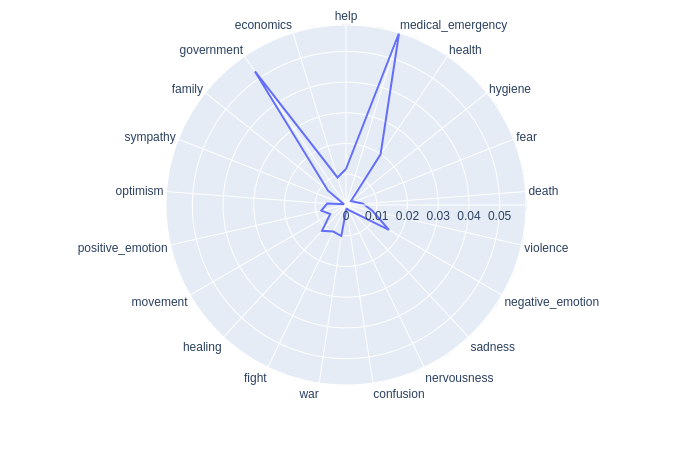} \\
		(d) Tamil Nadu & (e) Odisha & (f) Kerala\\
		
	\end{tabular}
	\caption{Empath analysis of tweets particular to Indian states: The most prominent categories across all states were medical\_emergency \& government.}
	\label{fig:empath-state-twitter}
\end{figure*}

\section{Analysis}
\subsection{Twitter Content}
\subsubsection{National Level}
Using Empath, we analysed the tweets collected in March 2020, along various psycho-linguistic attributes, as shown in Fig \ref{fig:empath-india}. The most common categories being discussed in the Tweets were government, health and medical emergency, which reflects that while discussing about the pandemic, the public is bringing the government in the discourse, be it referring to some government policy or some information released by the government. Another observation is that the negative emotion is fairly high amongst these tweets. A positive indicator is that the confusion level indicated by the Empath analysis is significantly low, while the frequency of linguistic features related to positive emotions, healing and optimism are much higher.\\
\begin{figure*}[ht!]
	\centering
	\begin{tabular}{ccc}
		\includegraphics[scale=0.25]{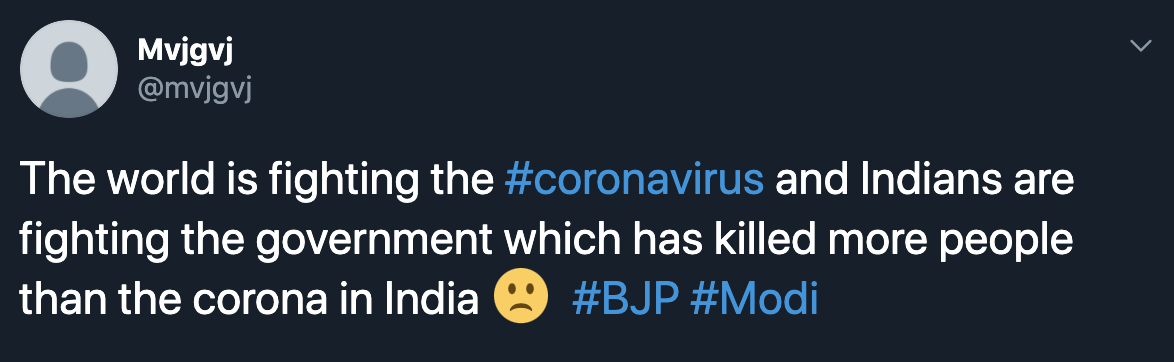} & 
		\includegraphics[scale=0.25]{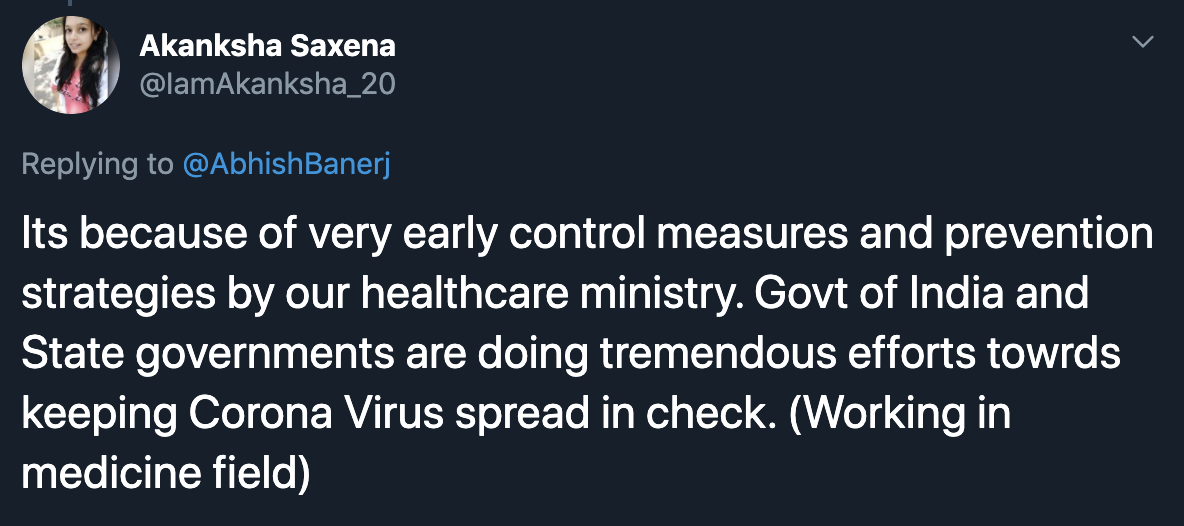} &
		\includegraphics[scale=0.25]{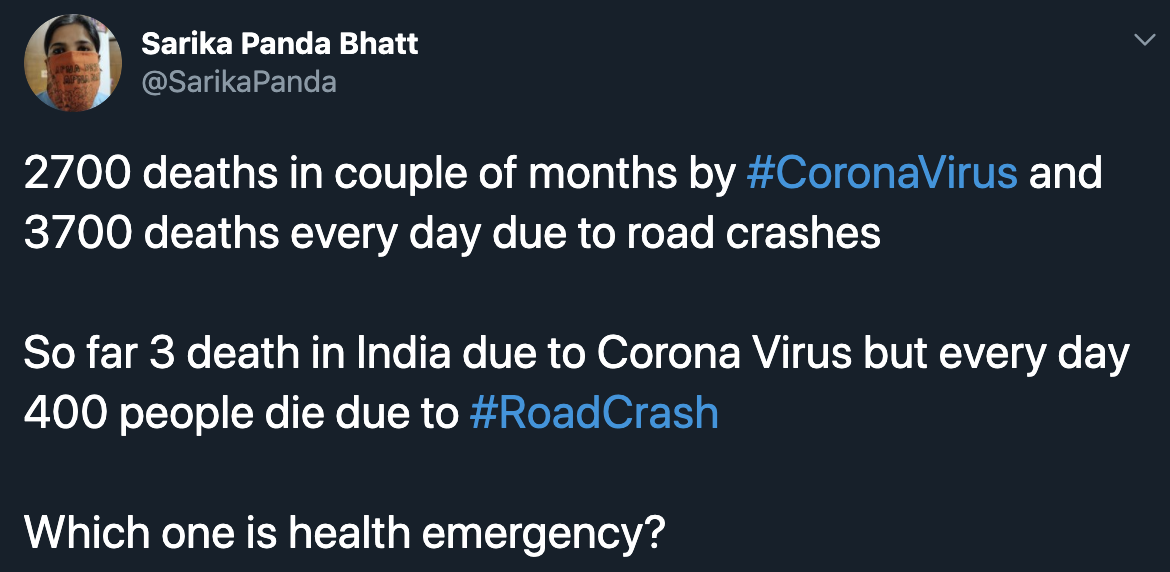} \\
		(a) Government-related & (b) Government-related & (c)Health emergency-related\\
		
	\end{tabular}
	\caption{Sample tweets belonging to the popular Empath categories.}
	\label{fig:sample-tweets}
\end{figure*}

\subsubsection{State Level}
We analysed the psycho-linguistic features of COVID-19 related discourses on Twitter at a state-level, as shown in Figure \ref{fig:empath-state-twitter}. 
It is interesting to note that although the magnitudes of different psycho-linguistic features vary across different states, their structure remains very similar.\\
We observed that while in some states like Delhi, Punjab, Odisha tweets talking about government related words were the most on Twitter, few states like Tamil Nadu and West Bengal talked more about medical emergency. Kerala, on the other hand had an equal frequency of words related to medical\_emergency and government. West Bengal and Kerala also have a higher frequency of words related to negative emotion compared to other states. \\

West Bengal showed higher levels of healing and positive emotions in the tweets. Interestingly, it also showed an even higher frequency of war or fight related words.

\subsection{Twitter Time Series}
% %%%%%%%%%%%%%%%%%%%%%%%%%%%%%%%%%%%%%%%%%%%%%%%%%%%%%%%%%%%%

% %%%%%%%%%%%%%%%% India Time Series Graphs %%%%%%%%%%%%%%%%%%
\begin{figure*}[ht!]
	\centering
	\begin{tabular}{ccc}
		\includegraphics[scale=0.3]{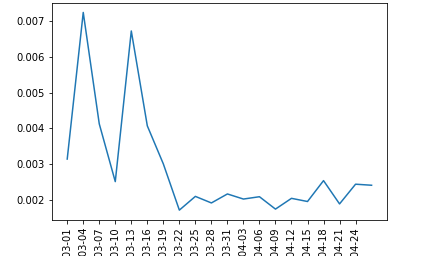} & 
		\includegraphics[scale=0.3]{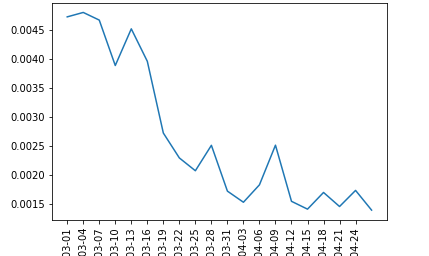} &
		\includegraphics[scale=0.3]{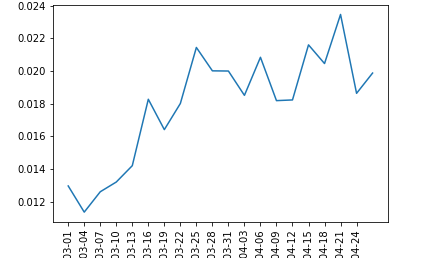} \\
		(a) Hygiene & (b) Nervousness & (c) Business \\
		\includegraphics[scale=0.3]{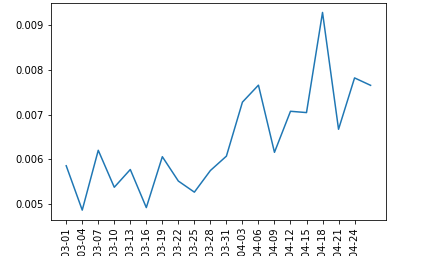} &
		\includegraphics[scale=0.3]{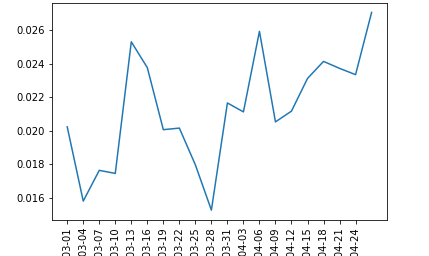} &
		\includegraphics[scale=0.3]{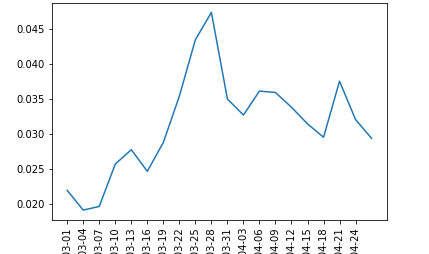} \\
		(d) Optimism & (e) Health & (f) Government\\

	\end{tabular}
	\caption{Time Series analysis of emotions in India related Tweets. Clear patterns of consistent decline/increase or peak-dip can be seen in the time series plots of various emotions, which correlate to on ground events related to COVID-19.}
	\label{fig:empath-time-series}
\end{figure*}

%%%%%%%%%%%%%%%%%%%%%%%%%%%%%%%%%%%%%%%%%%%%%%%%%%%%%%%%%%%%

We analysed the collected Twitter data in the Indian context over a period of 2 months (March and April 2020), by looking from the lens of various psycho-linguistic attributes as shown in Figure  \ref{fig:empath-time-series}. 
We observe that while the frequency of `hygiene' \& `nervousness' related words has decreased over time, since the start of COVID-19 crisis in India, words related to `business' \& `optimism' have become more frequent. The categories of `health' and `government' have been one of the most popular categories for the Indian twitter data, and while the presence of `health' category in tweets observes a sharp dip on March 28, `government' related words have a sharp rise on the same date.

\subsubsection{Relation of public sentiment to the rapidly changing on-ground situation}
We observed that the presence of optimism related keywords in the tweets has increased over time, with the highest frequency of optimistic words from 18th-21st April. It is interesting to note that on 18th April, due to the imposition of a nationwide lockdown, the time taken for doubling COVID-19 cases came down from every 3 days to every 8 days.\\
We also observe that the discussion regarding certain aspects of COVID-19 discourses on twitter have reduced over time, especially those related to `Hygiene' \& `movement', which became very popular near the time when the lockdown first got imposed, however, over time the frequency has declined possibly hinting at normalisation of certain aspects of the COVID-19 narrative on Indian twitter. \\
The frequency of nervousness related words has sharply declined over time, with the peak around the time when COVID-19 started becoming popular. We also observe that frequency of business-related words increase over time, with the peak being observed around 20th April, the day when the government allowed certain relaxation for shops, etc. to re-open up for the first time post the COVID-19 lockdown.\\
The frequency of `health' related tweets take a sharp dip near 28th March, wherein the frequency of tweets related to `government' observes its peak. It is interesting to note than on 28th March, India crossed a total number of 1,000 confirmed COVID-19 cases.
 We thus, observe that the rapidly evolving public sentiment are reflective of public's response to the on-ground n-CoV2019 situation and the government response.\\
 
% Similarly confusion, fear and sadness are some of the most popular categories being discussed throughout the time-series based collected data.

% We observed that confusion in the tweets peaked around second week March\footnote{March12th-14th}, when the stringent measures started getting imposed by the Indian state governments\footnote{On March 13th, Delhi government banned gatherings of more than 200 people}. Subsequently, confusion has been nearly consistently present till date, although it has reduced in magnitude than before.
% We also observed that as the spread of Coronavirus and its impact increased, discussions on Indian Twitter regarding economics, family, fear, etc peaked. The presence of linguistic features associated with medical emergency, fear, sadness, family, help spiked around the 13th-14th April, when India crossed 10,000 COVID-19 confirmed  cases and when the nationwide lockdown was further extended. We thus, observe that the rapidly evolving public sentiment are reflective of public's response to the on-ground coronavirus situation and the government response.\\

\subsubsection{The changing discourse around n-CoV2019}
As can be observed from Figure \ref{fig:empath-time-series} the discussion regarding certain aspects of COVID-19 discourses on twitter reduced over time, especially those related to Hygiene, movement and nervousness, while the discussions regarding business and optimism increased. We observe that hygiene-related COVID-19 discourses on Twitter became very popular near the time when the lockdown first got imposed. However, over time discussions regarding hygiene have reduced on Twitter, which might be due to it being normalised over time.\\

\begin{figure*}[ht!]
	\centering
	\begin{tabular}{ccc}
		\includegraphics[scale=0.2]{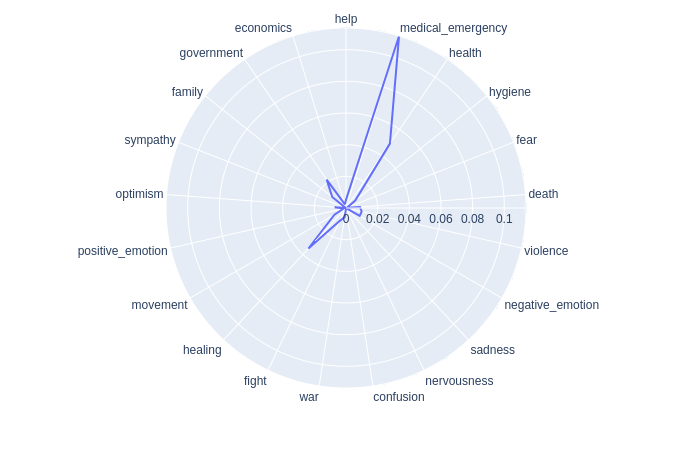} & 
		\includegraphics[scale=0.2]{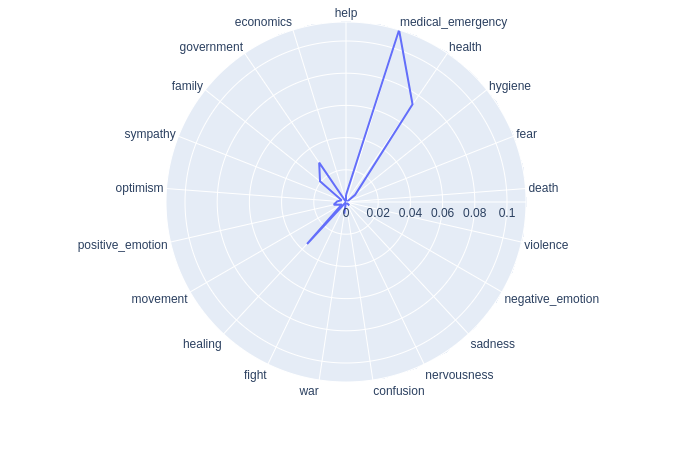} &
		\includegraphics[scale=0.2]{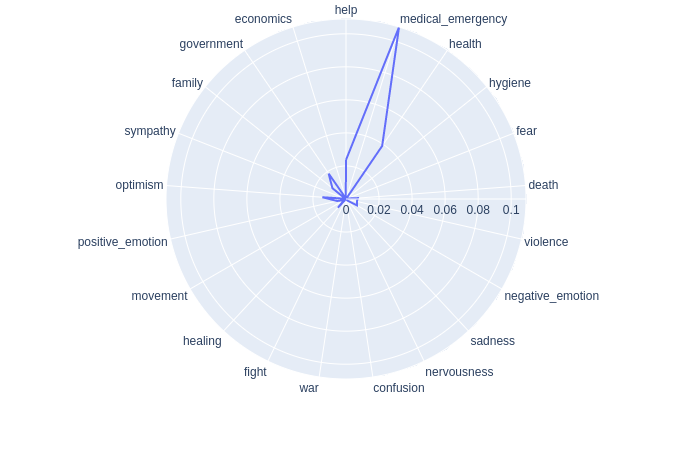} \\
		(a) Delhi & (b) West Bengal & (c) Punjab\\
		\includegraphics[scale=0.2]{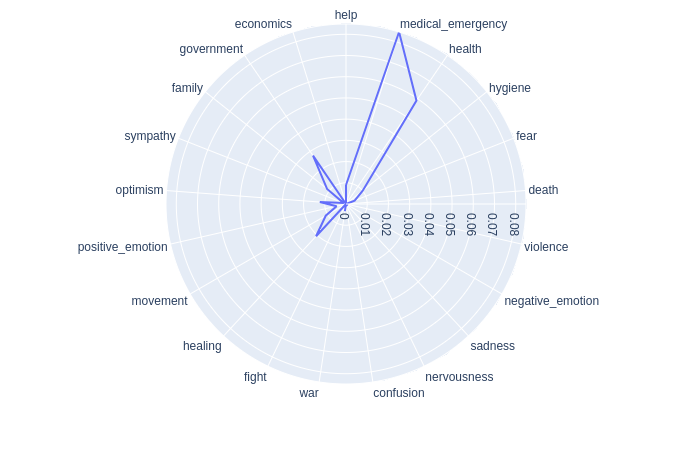} &
		\includegraphics[scale=0.2]{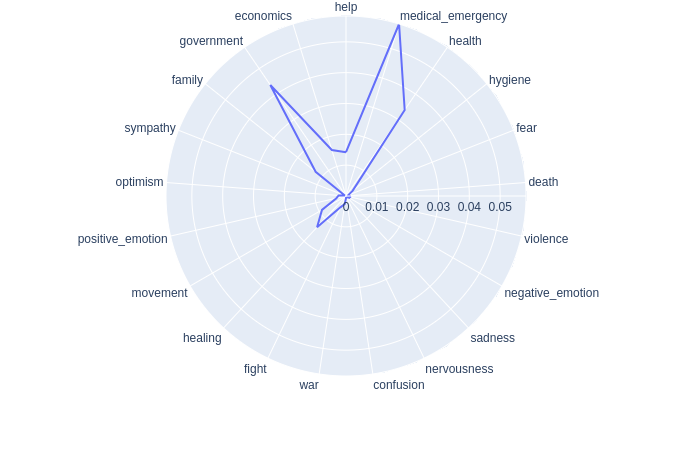} & \includegraphics[scale=0.2]{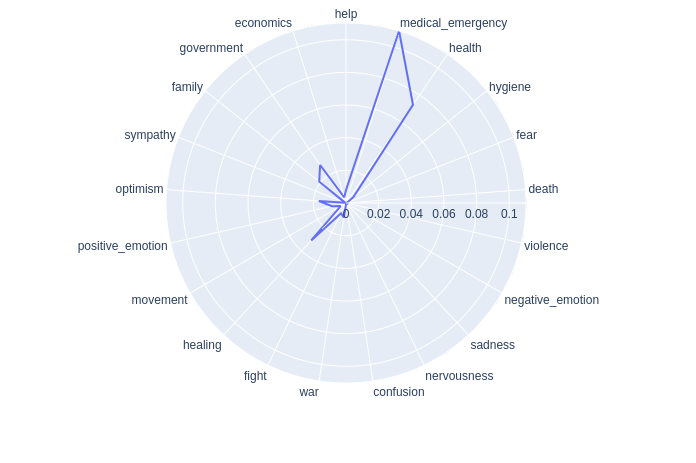}\\
		(d) Tamil Nadu & (e) Odisha & (f) Kerala \\
		
	\end{tabular}
	\caption{Empath analysis of Government Bulletins related to COVID-19 in India.}
	\label{fig:empath-govt-bulletins}
\end{figure*}

\subsection{Government Bulletins Content}
We observe that the government bulletins(as described in Section \ref{govt-bulletins}) shared by the governments of all the six states, as shown in Figure \ref{fig:empath-govt-bulletins} frequently use words related to medical emergency and health.
The government bulletins released on n-CoV2019, by the Delhi and West Bengal government\footnote{\url{https://github.com/datameet/covid-19-indian-state-reports}}, have a higher frequency of linguistic features related to the topic `healing' than other states. An interesting observation is that the state government bulletins in Odisha have a significantly higher inclination towards using words related to government, while for most other states, the primary focus is towards medical emergency. Also, all government bulletins show no `fear' or `confusion' related psycho-linguistic markers.

\subsection{Granger's Causality Analysis}
As a pre-requisite for studying causal mechanism between the time series on Delhi Bulletin and Delhi Tweets, both the sets of data were subjected to Augmented Dickey-Fuller (ADF) test of unit root (so as to see whether the series are stationary or not). The formulation adopted for the ADF test was:\\
\begin{equation}
\label{eqn:eq1}
\triangle{Y_{t}} = \tau Y_{t-1} + \alpha_{i}\sideset{}{}\sum_{i=1}^{m} \triangle{Y_{t-i}}+u_{t}
\end{equation}\\
where t stands for the time variable; $\triangle{}$ for the difference operator; and u$_{t}$ for the disturbance terms. The null and alternative hypotheses for the test are:\\
$\triangle{H_{0}}: \tau = 0$(meaning that the series possesses a unit root and is, therefore, non-stationary); \\
$\triangle{H_{1}}: \tau < 1$ (meaning that the series does not possess a unit root and is, therefore, stationary).\\
For both the time series, the test was performed \textit{at levels} as well as at \textit{first difference}; Table \ref{tab:ADF}. As per the table, value of the test statistic $\tau$ for the time series (at level) on \textit{Help} in respect of Delhi Bulletin was computed to be -1.596, which failed to reach the critical values ( -1.951 at 5\% and -2.623 at 1\% level of significance). Accordingly, we could not reject the null hypothesis of the presence of a \textit{unit root} in the series. In other words, the series on `Help’ was \textit{non-stationary}. However, at the first difference, the series was detected to be free from a unit root ($\tau$ = -6.081) and was, therefore, stationary in nature. A large majority of the time series on rest of the variables (except for Fear, Sadness,  Nervousness, Confusion, Fun, Positive Emotion and Economics, wherein stationarity was present at levels itself) of Delhi Bulletin showed this very type of behaviour. Notably, in respect of Delhi Tweet, the series on the entire set of variables were observed to be non-stationary at levels but stationary at the first difference. Consequently, for examining causality behaviour, we have uniformly considered the first-differenced series on all the variables in respect of both data sets. Here, we may mention that ADF test could not be performed on the variable \textit{Sympathy} in respect of the first set of data (because of the absence of variability) and was, therefore, left-out for the subsequent analysis on Granger’s causality.\\
For examining causality, each of the corresponding pairs of variables from the first-differenced data sets was subjected to the estimation of Equations 2 and 3:\\
\begin{equation}
\label{eqn:eq2}
Y_{t} = \alpha_{01} +\sideset{}{}\sum_{i=1}^{p}\alpha_{1i}Y_{t-i} + \sideset{}{}\sum_{i=1}^{p}\beta_{1i}X_{t-i}+u_{1t}\end{equation}\\
\begin{equation}
\label{eqn:eq3}
Y_{t} = \alpha_{02}+\sideset{}{}\sum_{i=1}^{p}\alpha_{2i}Y_{t-i} +u_{2t}\end{equation}\\
which were then compared for their predictive power through \textit{Wald’s test}. If Equation 2 turns out to be statistically superior to the Equation 1 (thus implying that current value of Y can be better predicted through its own past values as well as the past values of X than through the past values of Y alone), then we say X \textit{Granger causes} Y. The series Y and X were then interchanged and the process repeated so as to examine if Y \textit{Granger causes} X.\\
The optimum number p of lagged terms to be included was decided through \textit{min AIC} criterion which, in the present analysis, turned out to be 1, in general.\\

As per the results through the analysis (Table \ref{tab:GCT}), it was observed that in respect of the variable `Help’, the data set on Delhi Tweet (TWT) failed to induce any causality on the data set on Delhi Bulletin (BLT), because the F-value (= 1.194 at 1 \& 30 d.f.) associated with Wald's test turned out to be statistically non-significant (p = 0.2832). On interchanging the two data sets, the finding remained virtually similar, thus implying that no causal linkage could be detected between the two data sets with respect to `Help’.
\\

In respect of `Medical Emergency’, the data set on Tweet induced significant causality (at 5\% probability level) on such a data such on Bulletin. On interchanging the two data sets, strength of causality (from Bulletin to Tweet) became all the more robust (at 0.1\% probability level). We may thus say that although there was an indication of bi-directional (or, equivalently, feedback) causality between the two sets of data in respect of ‘Medical Emergency’, yet the strength of causality was more pronounced from Bulletin to Tweet. Bi- directional causality (at 5\% level) between the two sets of data was observed in respect of `Health’. Very strong causality (at < 0.1\% probability level) from Tweet to Bulletin was indicated in respect of the variables `Hygiene’, `Leisure’, `Fun’ and `Government’. Significant (at 5\% probability level) unidirectional causality (from Bulletin to Tweet) was also detected in respect of each of ‘Death’ and ‘War’. But for rest of the variables, causal linkages between the two sets of data could failed to be established. Thus, on the whole, direction and strength of causality between the two sets of data were peculiar to the variable under consideration.
\begin{table*}
\begin{threeparttable}
\centering
\caption{Results in respect of Augmented Dickey-Fuller test of Unit Root.}
\label{tab:ADF}
\begin{tabular}{|c|c|c|c|c|c|c|c|c|} 
\hline
                  & \multicolumn{4}{c|}{Bulletin}                                           & \multicolumn{4}{c|}{Tweet}                                               \\ 
\cline{2-9}
Variable          & \multicolumn{2}{c|}{At Levels} & \multicolumn{2}{c|}{First Differenced} & \multicolumn{2}{c|}{At Levels} & \multicolumn{2}{c|}{First Differenced}  \\ 
\cline{2-9}
                  & $\tau$  & Significance         & $\tau$  & Significance                 & $\tau$  & Significance         & $\tau$  & Significance                  \\ 
\hline
Help              & -1.5957 & NS                   & -6.0811 & **                           & -0.0196 & NS                   & -6.7529 & **                            \\
Medical Emergency & -0.7589 & NS                   & -4.2722 & **                           & -0.3510 & NS                   & -4.8404 & **                            \\
Health            & -0.708  & NS                   & -4.4264 & **                           & -0.1305 & NS                   & -5.3220 & **                            \\
Hygiene           & -1.0035 & NS                   & -3.9099 & **                           & -0.8163 & NS                   & -5.5162 & **                            \\
Fear              & -1.9712 & *                    & -4.0759 & **                           & -1.6783 & NS                   & -5.2862 & **                            \\
Death             & -1.0259 & NS                   & -5.5868 & **                           & -1.0432 & NS                   & -3.9519 & **                            \\
Negative Emotion  & -0.9951 & NS                   & -5.3831 & **                           & -1.1707 & NS                   & -4.9076 & **                            \\
Sadness           & -3.6667 & **                   & -6.8719 & **                           & -1.4853 & NS                   & -3.9579 & **                            \\
Nervousness       & -2.4119 & *                    & -4.1097 & **                           & -2.8012 & NS                   & -4.4155 & **                            \\
Confusion         & -2.2471 & *                    & -6.1171 & **                           & -0.3505 & NS                   & -5.1352 & **                            \\
War               & -0.7392 & NS                   & -4.1929 & **                           & -0.6573 & NS                   & -3.6262 & **                            \\
Fight             & -1.1623 & NS                   & -5.831  & **                           & -0.6296 & NS                   & -3.4056 & **                            \\
Healing           & -0.7563 & NS                   & -5.2625 & **                           & 0.0691  & NS                   & -4.0565 & **                            \\
Movement          & -0.5542 & NS                   & -3.4916 & **                           & -0.7894 & NS                   & -4.4361 & **                            \\
Leisure           & -1.4084 & NS                   & -4.4319 & **                           & -0.7163 & NS                   & -6.4206 & **                            \\
Fun               & -1.9783 & *                    & -5.1959 & **                           & -0.7581 & NS                   & -2.9320 & **                            \\
Positive Emotion  & -2.705  & **                   & -5.7106 & **                           & -0.9994 & NS                   & -3.8429 & **                            \\
Optimism          & -0.2408 & NS                   & -3.4171 & **                           & -0.7702 & NS                   & -3.8225 & **                            \\
Sympathy          & NC      & NC                   & NC      & NC                           & -1.6691 & NS                   & -5.8757 & **                            \\
Family            & -1.579  & NS                   & -4.7658 & **                           & -0.1853 & NS                   & -4.3916 & **                            \\
Government        & -0.6655 & NS                   & -4.9432 & **                           & 0.2586  & NS                   & -3.5532 & **                            \\
% Modi              & -2.4119 & *                    & -4.1097 & **                           & -0.8690 & NS                   & -5.5639 & **                            \\
Economics         & -2.2898 & *                    & -5.8496 & **                           & 0.0531  & NS                   & -4.1381 & **                            \\
Business          & -0.7952 & NS                   & -4.7221 & **                           & 0.2435  & NS                   & -2.9480 & **                            \\
Occupation        & -0.7609 & NS                   & -6.5216 & **                           & 0.0550  & NS                   & -4.3939 & **                            \\
\hline
\end{tabular}
\begin{tablenotes}\footnotesize
\item[*] \textbf{Critical Values of $\tau$:} -1.951 at 5\% and -2.623 at 1\% level of significance;\\***: Significant at 0.1\% probability level; **: Significant at 1\% probability level; *: Significant at 5\% probability level; NS: Non-significant; NC: Non-computable.
\end{tablenotes}
\end{threeparttable}
\end{table*}

\begin{table*}
\begin{threeparttable}
\centering
\caption{Results in respect of Granger's Causality Test.}
\label{tab:GCT}
\begin{tabular}{|c|c|c|c|c|c|} 
\hline
\textbf{Variable} & \textbf{Caused Set (Y)} & \textbf{Causal Set(X)} & \textbf{F-Value from Wald's Test}$^{\#}$ & \textbf{p-Value} & \textbf{Remark}$^{\$}$  \\ 
\hline
\multirow{2}{*}{Help} 
& Bulletin       & Tweet         & 1.194                    & 0.2832  & NS      \\ 
\cline{2-6}
                                  & Tweet          & Bulletin      & 1.455                    & 0.2372  & NS      \\ 
\hline
\multirow{2}{*}{Medical Emergency} & Bulletin       & Tweet         & 6.932                    & 0.0133  & *       \\ 
\cline{2-6}
                                  & Tweet          & Bulletin      & 16.673                   & 0.0003  & ***     \\ 
\hline
\multirow{2}{*}{Health}            & Bulletin       & Tweet         & 4.988                    & 0.0331  & *       \\ 
\cline{2-6}
                                  & Tweet          & Bulletin      & 5.99                     & 0.0205  & *       \\ 
\hline
\multirow{2}{*}{Hygiene}           & Bulletin       & Tweet         & 10.004                   & 0.0036  & **      \\ 
\cline{2-6}
                                  & Tweet          & Bulletin      & 2.71                     & 0.1102  & NS      \\ 
\hline
\multirow{2}{*}{Fear}              & Bulletin       & Tweet         & 0.014                    & 0.9078  & NS      \\ 
\cline{2-6}
                                  & Tweet          & Bulletin      & 0.115                    & 0.7364  & NS      \\ 
\hline
\multirow{2}{*}{Death}             & Bulletin       & Tweet         & 0.157                    & 0.6947  & NS      \\ 
\cline{2-6}
                                  & Tweet          & Bulletin      & 5.787                    & 0.0225  & *       \\ 
\hline
\multirow{2}{*}{Negative Emotion}  & Bulletin       & Tweet         & 0.036                    & 0.8505  & NS      \\ 
\cline{2-6}
                                  & Tweet          & Bulletin      & 2.558                    & 0.1202  & NS      \\ 
\hline
\multirow{2}{*}{Sadness}           & Bulletin       & Tweet         & 1.872                    & 0.1814  & NS      \\ 
\cline{2-6}
                                  & Tweet          & Bulletin      & 0.093                    & 0.7622  & NS      \\ 
\hline
\multirow{2}{*}{Nervousness}       & Bulletin       & Tweet         & 0.828                    & 0.37    & NS      \\ 
\cline{2-6}
                                  & Tweet          & Bulletin      & 0.034                    & 0.8542  & NS      \\ 
\hline
\multirow{2}{*}{Confusion}         & Bulletin       & Tweet         & 0.011                    & 0.9164  & NS      \\ 
\cline{2-6}
                                  & Tweet          & Bulletin      & 3.355                    & 0.077   & NS      \\ 
\hline
\multirow{2}{*}{War}               & Bulletin       & Tweet         & 0.031                    & 0.8616  & NS      \\ 
\cline{2-6}
                                  & Tweet          & Bulletin      & 6.425                    & 0.0167  & *       \\ 
\hline
\multirow{2}{*}{Fight}             & Bulletin       & Tweet         & 3.937                    & 0.0564  & NS      \\ 
\cline{2-6}
                                  & Tweet          & Bulletin      & 0.32                     & 0.5756  & NS      \\ 
\hline
\multirow{2}{*}{Healing}           & Bulletin       & Tweet         & 0.146                    & 0.7049  & NS      \\ 
\cline{2-6}
                                  & Tweet          & Bulletin      & 0.693                    & 0.4117  & NS      \\ 
\hline
\multirow{2}{*}{Movement}          & Bulletin       & Tweet         & 0.603                    & 0.4436  & NS      \\ 
\cline{2-6}
                                  & Tweet          & Bulletin      & 2.228                    & 0.146   & NS      \\ 
\hline
\multirow{2}{*}{Leisure}           & Bulletin       & Tweet         & 16.293                   & 0.0004  & ***     \\ 
\cline{2-6}
                                  & Tweet          & Bulletin      & 7.044                    & 0.0126  & *       \\ 
\hline
\multirow{2}{*}{Fun}               & Bulletin       & Tweet         & 36.026                   & 0.0001  & ***     \\ 
\cline{2-6}
                                  & Tweet          & Bulletin      & 5.561                    & 0.0251  & *       \\ 
\hline
\multirow{2}{*}{Positive Emotion}  & Bulletin       & Tweet         & 0.231                    & 0.6344  & NS      \\ 
\cline{2-6}
                                  & Tweet          & Bulletin      & 0.238                    & 0.6292  & NS      \\ 
\hline
\multirow{2}{*}{Optimism}          & Bulletin       & Tweet         & 1.734                    & 0.1978  & NS      \\ 
\cline{2-6}
                                  & Tweet          & Bulletin      & 1.263                    & 0.2701  & NS      \\ 
\hline
\multirow{2}{*}{Family}            & Bulletin       & Tweet         & 0.419                    & 0.5224  & NS      \\ 
\cline{2-6}
                                  & Tweet          & Bulletin      & 1.492                    & 0.2314  & NS      \\ 
\hline
\multirow{2}{*}{Government}        & Bulletin       & Tweet         & 9.097                    & 0.0052  & **      \\ 
\cline{2-6}
                                  & Tweet          & Bulletin      & 2.387                    & 0.1328  & NS      \\ 
\hline
% \multirow{2}{*}{Modi}              & Bulletin       & Tweet         & 7.594                    & 0.0099  & **      \\ 
% \cline{2-6}
%                                   & Tweet          & Bulletin      & 0.864                    & 0.3602  & NS      \\ 
% \hline
\multirow{2}{*}{Economics}         & Bulletin       & Tweet         & 0.071                    & 0.7916  & NS      \\ 
\cline{2-6}
                                  & Tweet          & Bulletin      & 0.195                    & 0.6622  & NS      \\ 
\hline
\multirow{2}{*}{Business}          & Bulletin       & Tweet         & 0.479                    & 0.4941  & NS      \\ 
\cline{2-6}
                                  & Tweet          & Bulletin      & 1.484                    & 0.2326  & NS      \\ 
\hline
\multirow{2}{*}{Occupation}        & Bulletin       & Tweet         & 2.573                    & 0.1192  & NS      \\ 
\cline{2-6}
                                  & Tweet          & Bulletin      & 3.497                    & 0.0713  & NS      \\
\hline
\end{tabular}
\begin{tablenotes}\footnotesize
\item[*] $^{\#:}$At 1 \& 30 d.f. $^{\$}$: ***: Significant at 0.1\% probability level; **: Significant at 1\% probability level; *: Significant at 5\% probability level; NS: Non-significant.
\end{tablenotes}
\end{threeparttable}
\end{table*}

\section{Conclusion}
We present novel dataset consisting of more than 5.6 million n-CoV2019 related Indian tweets, with special emphasis on the tweets related to each of the Indian states. We further analyse the tweets to find the important topics and psycho-linguistic features discussed and compare it both between various states, as well as using a time-series based approach. We further try to link the rapidly changing psycho-linguistic attributes of the public sentiment to the real-life on-ground situations arising due to COVID-19.  We further designed an interactive web portal  \href{http://covibes.tavlab.iiitd.edu.in/ }{COVibes}\footnote{\url{http://covibes.tavlab.iiitd.edu.in/ }}, which displays psychometric insights gained both at a national and state level, from the CoronaIndiaDataset. This dataset and analysis technique can be used for further research into understanding the public perceptions and taking more effective policy decisions. We restricted our work to the analysis of tweets as well as government bulletins in English. Future work in this direction can be done to increase the scope of the analysis to various Indian languages.

% \newpage
% \bibliographystyle{unsrt}
% \bibliographystyle{these}
\bibliographystyle{acm}
% \bibliographystyle{elsart-harv}
% \bibliographystyle{ieeetr}
% \newpage
\bibliography{references}

\begin{thebibliography}{10}

\bibitem{Alhajji2019}
{\sc Alhajji, M., Al~Khalifah, A., Aljubran, M., and Alkhalifah, M.}
\newblock Sentiment analysis of tweets in saudi arabia regarding governmental
  preventive measures to contain covid-19.

\bibitem{Chen2020}
{\sc Chen, E., Lerman, K., and Ferrara, E.}
\newblock Covid-19: The first public coronavirus twitter dataset.
\newblock {\em arXiv preprint arXiv:2003.07372\/} (2020).

\bibitem{cinelli2020covid}
{\sc Cinelli, M., Quattrociocchi, W., Galeazzi, A., Valensise, C.~M., Brugnoli,
  E., Schmidt, A.~L., Zola, P., Zollo, F., and Scala, A.}
\newblock The covid-19 social media infodemic.
\newblock {\em arXiv preprint arXiv:2003.05004\/} (2020).

\bibitem{Empath-Fast2016}
{\sc Fast, E., Chen, B., and Bernstein, M.~S.}
\newblock {Empath: Understanding topic signals in large-scale text}.
\newblock {\em Conference on Human Factors in Computing Systems -
  Proceedings\/} (2016), 4647--4657.

\bibitem{haouari2020ArCov-19}
{\sc Haouari, F., Hasanain, M., Suwaileh, R., and Elsayed, T.}
\newblock Arcov-19: The first arabic covid-19 twitter dataset with propagation
  networks.
\newblock {\em arXiv preprint arXiv:2004.05861\/} (2020).

\bibitem{Hou2020}
{\sc Hou, Z., Du, F., Jiang, H., Zhou, X., and Lin, L.}
\newblock {Assessment of Public Attention, Risk Perception, Emotional and
  Behavioural Responses to the COVID-19 Outbreak: Social Media Surveillance in
  China}.
\newblock {\em SSRN Electronic Journal\/} (2020).

\bibitem{jain2015effective}
{\sc Jain, V.~K., and Kumar, S.}
\newblock An effective approach to track levels of influenza-a (h1n1) pandemic
  in india using twitter.
\newblock {\em Procedia Computer Science 70\/} (2015), 801--807.

\bibitem{Kayes2020}
{\sc Kayes, A., Islam, M.~S., Watters, P.~A., Ng, A., and Kayesh, H.}
\newblock Automated measurement of attitudes towards social distancing using
  social media: A covid-19 case study.

\bibitem{kim2016topic}
{\sc Kim, E. H.-J., Jeong, Y.~K., Kim, Y., Kang, K.~Y., and Song, M.}
\newblock Topic-based content and sentiment analysis of ebola virus on twitter
  and in the news.
\newblock {\em Journal of Information Science 42}, 6 (2016), 763--781.

\bibitem{lazard2015detecting}
{\sc Lazard, A.~J., Scheinfeld, E., Bernhardt, J.~M., Wilcox, G.~B., and Suran,
  M.}
\newblock Detecting themes of public concern: a text mining analysis of the
  centers for disease control and prevention's ebola live twitter chat.
\newblock {\em American journal of infection control 43}, 10 (2015),
  1109--1111.

\bibitem{Li2020b}
{\sc Li, L., Zhang, Q., Wang, X., Zhang, J., Wang, T., Gao, T.-L., Duan, W.,
  Tsoi, K. K.-f., and Wang, F.-Y.}
\newblock {Characterizing the Propagation of Situational Information in Social
  Media During COVID-19 Epidemic: A Case Study on Weibo}.
\newblock {\em IEEE Transactions on Computational Social Systems 7}, 2 (2020),
  1--7.

\bibitem{Li2020}
{\sc Li, S., Wang, Y., Xue, J., Zhao, N., and Zhu, T.}
\newblock {The impact of covid-19 epidemic declaration on psychological
  consequences: A study on active weibo users}.
\newblock {\em International Journal of Environmental Research and Public
  Health 17}, 6 (2020).

\bibitem{liu2012sentiment}
{\sc Liu, B.}
\newblock Sentiment analysis and opinion mining.
\newblock {\em Synthesis lectures on human language technologies 5}, 1 (2012),
  1--167.

\bibitem{Ritterman2009}
{\sc Ritterman, J., Osborne, M., and Klein, E.}
\newblock Using prediction markets and twitter to predict a swine flu pandemic.
\newblock In {\em 1st international workshop on mining social media\/} (2009),
  vol.~9, pp.~9--17.

\bibitem{Schild2020}
{\sc Schild, L., Ling, C., Blackburn, J., Stringhini, G., Zhang, Y., and
  Zannettou, S.}
\newblock {"Go eat a bat, Chang!": An Early Look on the Emergence of Sinophobic
  Behavior on Web Communities in the Face of COVID-19}.

\bibitem{signorini2011use}
{\sc Signorini, A., Segre, A.~M., and Polgreen, P.~M.}
\newblock The use of twitter to track levels of disease activity and public
  concern in the us during the influenza a h1n1 pandemic.
\newblock {\em PloS one 6}, 5 (2011).

\bibitem{wong2017local}
{\sc Wong, R., Harris, J.~K., Staub, M., and Bernhardt, J.~M.}
\newblock Local health departments tweeting about ebola: characteristics and
  messaging.
\newblock {\em Journal of Public Health Management and Practice 23}, 2 (2017),
  e16--e24.

\bibitem{Zhao2020}
{\sc Zhao, Y., and Xu, H.}
\newblock Chinese public attention to covid-19 epidemic: Based on social media.
\newblock {\em medRxiv\/} (2020).

\end{thebibliography}
% \endinput
\clearpage
\appendix
% \title{Appendix:}
\section{Appendix A: Hashtags used to collect Tweets}

\subsection{Content-based query:}
We manually curated the following list of hashtags, based on trending and popular hashtags used at India level as well as each of the states:\\
\#CoronainMaharastra, \#CoronainKolkata, \#CoronavirusReachesDelhi, \#CoronainDelhi, \#CoronainBengal, \#CoronaDelhi, \#COVID19-Delhi, \#COVIDDelhi, \#coronavirusindelhi, \#CoronainPunjab, \#CoronaPunjab, \#COVID19Punjab, \#COVIDPunjab, \#CoronaBengaluru, \#CoronaKarnataka, \#CoronainKarnataka, \#COVID19Bangalore, \#CoronaBangalore, \#CoronavirusBangalore, \#CoronavirusinBangalore, \#CoronaInKerala, \#CoronaVirusAgra, \#CoronaVirusUp, \#Coronaintamilnadu, \#coronatamilnadu, \#COVIDLadakh, \#coronagurgaon, \#coronatelangana, \#COVID19telangana, \#COVIDtelangana, \#coronahyderabad, \#COVIDhyderabad, \#COVID19hyderabad, \#coronavirusinhyderabad, \#coronavirushyderabad, \#CoronainUP, \#CoronainNoida, \#COVID19Noida, \#CoronaRajasthan, \#CoronainRajasthan, \#CoronaBihar, \#CoronainBihar, \#coronauttarkhand, \#coronaOdisha, \#coronaOdisha, \#COVID19Odisha, \#COVIDOdisha, \#COVIDAndhra-Pradesh, \#covidinindia, \#coronavirusindia, \#coronainindia, \#coronaindia, \#covidindia, \#covid19india. \\

Based on the above list, we identified a certain set of patters with region names or aliases. We used the following prefixes to add before each region name or it's aliases (`term <region>') to run the queries - `corona ', `lockdown ', `coronain ', `covidin ', `stayathome ', `covid ', `coronavirus '. In addition, we use these prefixes before region names to create hashtags (\#prefix<region>) - \#covid, \#corona, \#coronavirus, \#covidin, \#coronain, \#coronavirusin, \#covid19, \#covid2019, \#<region>fightscorona.

% \subsection{Location-based query:}

% The hashtags and keywords used to collect the twitter data were as below:\\
% \#COVID19, `CoronaVirusUpdates', `coronavirus', `corona virus outbreak', `corona wuhan', \#Coronavirus, \#NCOV19, \#CoronavirusOutbreak, \#coronaviruschina, \#coronavirus, `COVID19'

\end{document}